\documentclass[journal]{IEEEtran}
\usepackage{graphicx}
\usepackage{booktabs}
\usepackage{amsmath}
\usepackage{amssymb}
\usepackage{cite}
\usepackage{xcolor}
\usepackage[normalem]{ulem}
\usepackage[colorlinks=true,citecolor=blue,linkcolor=blue,urlcolor=blue]{hyperref}

\title{Frequency–Structure Complementary Modeling for Hyperspectral Image Fusion}
\author{
Qiya~Song, Siyu~Tian, Shutao~Li, Renwei~Dian, Long~Sun, and Jinyang~Liu
\thanks{The authors are with the School of Information Science and Engineering, Hunan Normal University, Changsha 410081, China.}
\thanks{Corresponding author: Renwei Dian (e-mail: drw@hnu.edu.cn).}
}

\date{}

\begin{document}  
\maketitle
\begin{abstract}
Hyperspectral image (HSI) and multispectral image (MSI) fusion seeks to reconstruct a high-spatial-resolution HSI from a low-spatial-resolution HSI and a high-spatial-resolution MSI by exploiting their complementary spectral and spatial information. Existing methods mainly employ convolutional aggregation or attention-based spatial-spectral interaction. However, conventional feature-domain attention does not explicitly characterize variation patterns along spatial-token and learned latent-channel coordinates, while spatially shared local operators lack the flexibility to adapt neighborhood aggregation to heterogeneous structures. To address these limitations, we propose a Frequency--Structure Complementary Modeling Network (FSCM-Net), which couples frequency-aware dependency learning with structure-adaptive local refinement. Specifically, a Spatial-Spectral Frequency Calibration Attention (SSFCA) module constructs spatial-token affinities from fourier-interacted query-key representations and builds latent-channel affinities after input-conditioned frequency filtering along learned channel coordinates. The latent-channel response is then calibrated by a learned shared gain and combined with the spatial-token response. Complementarily, a Structure-Coherent Dynamic Feature Fusion (SC-DFF) module combines regularized structure-tensor cues from MSI features with context-modulated HSI detail representations to generate channel- and position-dependent aggregation kernels. This design makes local HSI refinement responsive to smooth regions, edges, and textures while maintaining spectral fidelity. The two modules are progressively integrated within a multiscale encoder-decoder to coordinate representation-level interaction and structure-adaptive reconstruction. Experiments on two simulated benchmarks and one real remote-sensing dataset demonstrate favorable performance against representative fusion methods in spatial detail reconstruction and spectral preservation. 
\end{abstract}

\section{Introduction}
\label{sec:introduction}
Hyperspectral imaging provides dense spectral measurements across numerous contiguous bands and has been widely used in land-cover classification~\cite{wang2023spectral}, target detection~\cite{yang2024variational}, agricultural and forestry monitoring~\cite{adao2017hyperspectral}, environmental monitoring~\cite{stuart2019hyperspectral}, and geological exploration~\cite{van2012multi}. However, physical constraints on imaging sensors impose an inherent tradeoff between spatial and spectral resolutions. Hyperspectral and multispectral image fusion seeks to reconstruct a high-spatial-resolution hyperspectral image (HR-HSI) from a low-spatial-resolution hyperspectral image (LR-HSI) and a high-resolution multispectral image (HR-MSI), aiming to recover fine spatial details while preserving spectral information~\cite{vivone2023multispectral}.

Early HSI-MSI fusion methods mainly relied on model-driven formulations, including coupled matrix factorization, statistical estimation, and subspace regularization~\cite{yokoya2011coupled,wei2015fast,simoes2014convex}. Although these methods provide interpretable reconstruction procedures, their performance often depends on handcrafted priors and accurate degradation models. To reduce reliance on manually designed reconstruction priors, deep-learning-based approaches learn nonlinear mappings from multimodal observations, improving the ability to represent complex spatial-spectral relationships~\cite{dian2018deep,hu2021hyperspectral}. Within this learning-based framework, convolutional neural networks efficiently capture local spatial-spectral patterns, while Transformer-based methods further extend feature interaction through content-dependent attention~\cite{hu2022fusformer,deng2023psrt,sun2024mdc,chen2024cyclic}. Despite these advances, effective fusion still requires improvements in two closely related aspects: representation-level interaction for exchanging complementary spatial-spectral information and reconstruction-level aggregation for recovering locally varying structures.

At the representation level, attention mechanisms facilitate information exchange by constructing dependencies primarily from feature-domain similarities. Volumetric attention further combines spatial-token and channel relationships to achieve more comprehensive interaction within hyperspectral feature representations~\cite{yu2025volformer}. However, such dependencies remain primarily defined in the original feature domain, without explicitly characterizing frequency-domain variation patterns along spatial-token and learned latent-channel coordinates. Frequency-domain processing provides a complementary representation of feature variations and has demonstrated its effectiveness in image restoration and hyperspectral fusion ~\cite{kong2023efficient,liang2024fourier,sun2024advancing}. These findings suggest that incorporating frequency information into spatial-spectral dependency learning can enrich the interactions established in the feature domain. 

At the reconstruction level, the complementary information obtained through feature interaction must be aggregated according to local image structures. Conventional convolutional operators
apply spatially shared neighborhood transformations, limiting their ability
to adjust aggregation weights to local edges and textures. Structure-tensor analysis provides explicit descriptors of local gradient organization and structural coherence~\cite{weickert1999coherence}, whereas dynamic filter networks generate input-conditioned filters for adaptive local processing~\cite{jia2016dynamic}, and dynamic convolution combines multiple kernels using input-dependent attention~\cite{chen2020dynamic}.
Combining these mechanisms can therefore translate structural cues into
adaptive aggregation, complementing frequency-aware interaction with
structure-sensitive local refinement.

Building on these observations, we propose a Frequency--Structure Complementary Modeling Network (FSCM-Net) for HSI-MSI fusion. FSCM-Net couples frequency-aware dependency learning with structure-adaptive local refinement within a unified multiscale encoder-decoder. Rather than treating representation interaction and local reconstruction as independent processes, the proposed framework progressively integrates them across multiple scales, allowing frequency-enhanced multimodal representations and structure-refined HSI features to complement each other. Specifically, we develop a Spatial–Spectral Frequency Calibration Attention (SSFCA) module to enrich spatial-token and latent-channel dependencies. SSFCA constructs spatial-token affinities from fourier-interacted query–key representations and builds latent-channel affinities after input-conditioned frequency filtering along learned channel coordinates. The latent-channel response is then calibrated by a learned shared gain and added to the spatial-token response, integrating the two complementary representations. The input-dependent modeling is provided by the attention affinities and dynamic latent spectral-frequency filtering, while the shared gain adjusts the relative contribution of the latent-channel branch. Complementarily, we design a Structure-Coherent Dynamic Feature Fusion (SC-DFF) module for structure-adaptive local HSI refinement. SC-DFF combines regularized structure-tensor cues extracted from MSI features with context-modulated HSI detail representations to predict channel- and position-dependent aggregation kernels. The MSI structural cues and HSI detail context determine the aggregation weights, whereas the complete HSI features provide the values being aggregated. Consequently, the local refinement can accommodate structural variations without directly replacing HSI feature values with MSI features.

The main contributions of this work are summarized as follows:
\begin{itemize}
    \item We propose a Frequency--Structure Complementary Modeling Network (FSCM-Net) to address insufficient frequency-aware dependency learning and limited structural adaptability by integrating frequency-aware interaction with structure-adaptive local refinement in a unified multiscale framework.

    \item We develop Spatial--Spectral Frequency Calibration Attention (SSFCA) to enhance spatial-token and latent-channel dependencies through Fourier-interacted query--key representations, input-conditioned latent-channel frequency filtering, and shared branch calibration.

    \item We design Structure-Coherent Dynamic Feature Fusion (SC-DFF) to improve local structural adaptability by combining regularized MSI structure-tensor cues with context-modulated HSI details to generate channel- and position-dependent aggregation kernels.
    \item Experiments on two simulated benchmarks and one real remote-sensing dataset demonstrate favorable spatial detail reconstruction and spectral preservation. 

\end{itemize}
The remainder of this paper is organized as follows. Section II reviews related work. Section III presents the proposed FSCM-Net. Section IV reports the experimental results and analyses. Finally, Section V concludes this paper.

\section{Related Work}
\label{sec:related_work}

\subsection{Hyperspectral Image Fusion}
Early HSI-MSI fusion methods reconstructed HR-HSIs by incorporating observation models and handcrafted image priors.  HySure~\cite{simoes2014convex} combines spectral subspace modeling with spatial regularization to constrain the reconstruction, while Wei et al.~\cite{wei2015fast} reduce the computational cost of model-based fusion by solving a Sylvester equation. Although these methods provide interpretable reconstruction procedures, they can be sensitive to mismatches between assumed priors or degradation models and actual sensor observations. CNN-Fus~\cite{dian2020regularizing} bridges model-based optimization and learned image priors by incorporating a pretrained CNN denoiser into subspace coefficient estimation. Deep-learning-based methods improve fusion by learning complementary representations directly from data. The two-branch CNN proposed by Yang et al.~\cite{yang2018hyperspectral} extracts spatial and spectral information separately, while GuidedNet~\cite{ran2023guidednet} employs multiscale MSI guidance to improve residual-detail reconstruction. Dian et al.~\cite{dian2018deep} learn image priors through CNN-based residual learning for hyperspectral sharpening, whereas HSRnet~\cite{hu2021hyperspectral} combines spatiospectral attention and pixelShuffle operations to recover spatial details while preserving spectral information. Nevertheless, conventional convolutional kernels apply spatially shared neighborhood transformations and have limited flexibility in adapting their aggregation weights to heterogeneous local structures. Tensor-based deep models further exploit the multidimensional structure of hyperspectral data. UDTN~\cite{yang2023unsupervised} couples tensor filtering with co-attention to learn shared latent tensor representations without HR-HSI supervision. DTDNML~\cite{wang2024unsupervised} integrates deep Tucker decomposition, spatial--spectral attention, and manifold constraints for unsupervised blind fusion. Transformer-based fusion methods further extend multimodal feature interaction. Fusformer~\cite{hu2022fusformer} introduces nonlocal attention to model contextual dependencies, whereas PSRT~\cite{deng2023psrt} promotes cross-window communication through pyramid shuffle-and-reshuffle operations. SST~\cite{chen2023spectral} and
SSTF-Unet~\cite{liu2023sstf} further explore complementary
spatial and spectral dependencies through dual-branch
Transformer designs. These approaches improve spatial--spectral information exchange, but their dependencies are mainly constructed from feature-domain similarities. Moreover, enhanced feature interaction does not directly specify how complementary information should be converted into structure-adaptive local HSI reconstruction.
\begin{figure*}[t]
    \centering
    \includegraphics[width=0.98\textwidth]{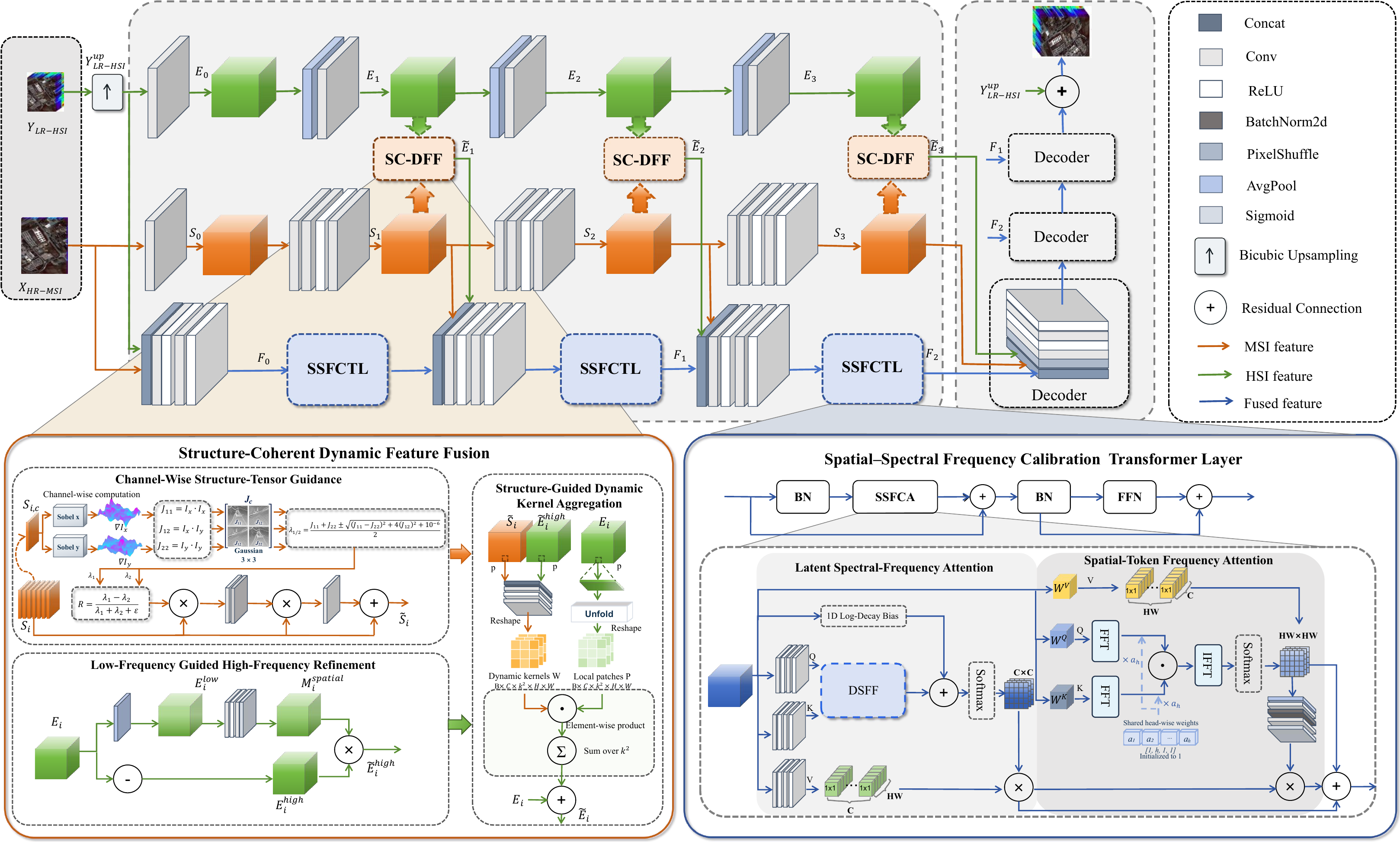}
\caption{
Overall architecture of FSCM-Net, which couples frequency-aware dependency learning with structure-adaptive local refinement within a multiscale encoder--decoder. The network comprises an MSI spatial branch, a structure-enhanced HSI branch, and a progressive fusion branch. SC-DFF combines regularized MSI structure-tensor cues with context-modulated HSI details to generate channel- and position-dependent aggregation kernels. SSFCA constructs spatial-token and latent-channel affinities through fourier interaction and input-conditioned frequency filtering, respectively. Their progressive integration provides complementary representations for HR-HSI reconstruction.
}
    \label{fig:overall_architecture}
\end{figure*}
\subsection{Spatial-Spectral Interaction and Adaptive Reconstruction}
Attention mechanisms improve image restoration by modeling content-dependent relationships beyond fixed convolutional neighborhoods. SwinIR~\cite{liang2021swinir} employs shifted-window spatial attention to exchange information within and across local regions, whereas Restormer~\cite{zamir2022restormer} constructs transposed channel attention to reduce the computational cost of processing high-resolution images. Although these approaches effectively model spatial-token or channel dependencies, they generally emphasize one dimension at a time. For single-HSI restoration, VolFormer~\cite{yu2025volformer} extends conventional attention through volumetric self-attention, jointly constructing spatial and spectral dependencies within hyperspectral feature representations. Its spectral positional prior further incorporates spectral-order information into channel interaction. Nevertheless, the resulting dependencies are still primarily constructed from feature-domain similarities. Frequency-domain modeling provides a complementary perspective for representation interaction. ITU-Net~\cite{sun2024advancing} integrates information-transfer-guided linearized attention and frequency-domain feedforward learning within an unfolding framework. FeINFN~\cite{liang2024fourier} improves high-frequency reconstruction through spatial-frequency implicit representations, while SSDAN~\cite{song2026s} combines global-local frequency fusion with differential feature complementarity to recover spatial structures and textures.  However, frequency-domain enhancement alone does not determine how local HSI neighborhoods should be aggregated under spatially varying structures. Structure-guided reconstruction provides another means of improving spatial-detail recovery. PSRF-DiffNet~\cite{qu2025progressive} jointly addresses registration and fusion by using an advanced structure tensor to enhance MSI details and guided filtering to transfer spatial information. Such methods demonstrate that local reconstruction can benefit from explicit structural guidance. Nevertheless, existing approaches do not explicitly combine MSI structural organization with the detail context already present in HSI representations to generate channel- and position-dependent neighborhood kernels.

\section{Proposed Method}
\label{sec:proposed_method}
\subsection{Overall Architecture}
\label{subsec:overall_architecture}

As illustrated in Fig.~\ref{fig:overall_architecture}, the proposed Frequency--Structure Complementary Modeling Network (FSCM-Net) couples frequency-aware dependency learning with structure-adaptive local refinement within a U-Net-style multiscale encoder--decoder. The framework consists of an MSI spatial branch, a structure-enhanced HSI branch, and a progressive fusion branch. 

Given an HR-MSI $\mathbf X\in\mathbb R^{C_m\times H\times W}$ and an LR-HSI $\mathbf Y\in\mathbb R^{C_h\times h\times w}$, the LR-HSI is first bicubically upsampled to the target spatial resolution, yielding $\mathbf Y_{\mathrm{up}}\in\mathbb R^{C_h\times H\times W}$. The MSI and HSI branches extract hierarchical modality-specific features at multiple spatial scales. In the HSI branch, Structure-Coherent Dynamic Feature Fusion (SC-DFF) combines regularized structure-tensor cues from MSI features with context-modulated HSI detail representations to predict channel- and position-dependent aggregation kernels. These kernels adaptively aggregate local HSI neighborhoods according to structural variations while retaining HSI features as the values being aggregated. The resulting structure-refined HSI features are subsequently incorporated into the progressive fusion branch through Spatial--Spectral Frequency Calibration Attention (SSFCA). SSFCA constructs spatial-token affinities from fourier-interacted query--key representations and latent-channel affinities after input-conditioned frequency filtering along learned channel coordinates. Specifically, the initial fusion stage applies SSFCA to shallow MSI and HSI features. At subsequent stages, the corresponding MSI features and SC-DFF-refined HSI features are concatenated with the preceding fused representation and processed by SSFCA for frequency-aware dependency learning. In this manner, structure-adaptive local refinement precedes frequency-aware interaction at subsequent fusion stages, enabling complementary information to be progressively integrated across scales.

At the bottleneck, the MSI, refined HSI, and fused representations are aggregated and passed to the decoder. Hierarchical upsampling and skip connections progressively restore the spatial resolution, producing the full-resolution feature $\mathbf F_{\mathrm{dec}}$. A reconstruction mapping $\mathcal R$ then predicts a residual that is added to the bicubically upsampled LR-HSI:
\begin{equation}
\mathbf Z
=
\mathbf Y_{\mathrm{up}}
+
\mathcal R(\mathbf F_{\mathrm{dec}}),
\label{eq:pm_reconstruction}
\end{equation}
where $\mathbf Z\in\mathbb R^{C_h\times H\times W}$ denotes the reconstructed HR-HSI. Through joint optimization under the reconstruction objective, FSCM-Net coordinates representation-level interaction and structure-adaptive local refinement to support spatial detail recovery and spectral preservation.

\subsection{Structure-Coherent Dynamic Feature Fusion}
\label{subsec:scdff}

Following multiscale feature extraction in the MSI and HSI branches, Structure-Coherent Dynamic Feature Fusion (SC-DFF) performs structure-adaptive local refinement of HSI features before the subsequent frequency-aware fusion stages, as illustrated in Fig.~\ref{fig:overall_architecture}. It predicts channel- and position-dependent neighborhood aggregation kernels using regularized structure-tensor cues from MSI features and context-modulated HSI detail representations. The MSI pathway supplies structural guidance, whereas the HSI pathway provides detail context from the current hyperspectral representation. Their joint conditioning determines how local HSI features are aggregated without directly replacing HSI values with MSI features.

At the $i$-th scale, let $\mathbf E_i$ and $\mathbf S_i$ denote the HSI feature and MSI guidance feature, respectively, both with dimensions $C_i\times H_i\times W_i$. SC-DFF can be summarized as
\begin{equation}
\begin{aligned}
\mathbf K_i
&=
\mathcal K
\left(
[
\widetilde{\mathbf E}_i^{\mathrm{high}},
\widetilde{\mathbf S}_i
]
\right),\\
\widetilde{\mathbf E}_i
&=
\operatorname{SC\text{-}DFF}
(\mathbf E_i,\mathbf S_i)\\
&=
\mathbf E_i
+
\mathcal A
(\mathbf E_i,\mathbf K_i),
\end{aligned}
\label{eq:pm_scdff_output}
\end{equation}
where $[\cdot,\cdot]$ denotes channel-wise concatenation, $\widetilde{\mathbf S}_i$ denotes the structure-encoded MSI representation, and $\widetilde{\mathbf E}_i^{\mathrm{high}}$ is the context-modulated HSI detail representation, $\mathcal K(\cdot)$ predicts spatially adaptive kernels, and $\mathcal A(\cdot)$ performs local neighborhood aggregation.

\subsubsection{Channel-Wise Structure-Tensor Guidance}

To encode second-order local gradient statistics, we construct a structure tensor independently for each MSI feature channel. Its diagonal and cross terms jointly describe local gradient variation. A regularized tensor-contrast response is then used as an intermediate input to learned MSI feature modulation.

For the $c$-th channel of the MSI feature $\mathbf S_i$, the horizontal and vertical gradient responses, denoted by $\nabla I_{x,c}$ and $\nabla I_{y,c}$ in Fig.~\ref{fig:overall_architecture}, are computed using Sobel operators:
\begin{equation}
\nabla I_{x,c}
=
K_x*S_{i,c},
\qquad
\nabla I_{y,c}
=
K_y*S_{i,c},
\label{eq:pm_sobel}
\end{equation}
where $K_x$ and $K_y$ denote the horizontal and vertical Sobel kernels, respectively, and $*$ denotes spatial convolution.

The corresponding second-order structure tensor is defined as
\begin{equation}
\mathbf J_c
=
\begin{bmatrix}
K_G*(\nabla I_{x,c}^{2})
&
K_G*(\nabla I_{x,c}\nabla I_{y,c})
\\
K_G*(\nabla I_{x,c}\nabla I_{y,c})
&
K_G*(\nabla I_{y,c}^{2})
\end{bmatrix},
\label{eq:pm_structure_tensor}
\end{equation}
where
\begin{equation}
K_G
=
\frac{1}{16}
[1,2,1]^\top
[1,2,1]
\label{eq:pm_gaussian_kernel}
\end{equation}
denotes a Gaussian smoothing kernel.

Let $J_{11,c}$, $J_{12,c}$, and $J_{22,c}$ denote the independent tensor entries. Following the implementation, we compute a regularized discriminant response:
\begin{equation}
\Delta_{i,c}
=
\sqrt{
(J_{11,c}-J_{22,c})^2
+
4J_{12,c}^{2}
+
\epsilon
},
\label{eq:pm_eigen_gap}
\end{equation}
The implementation represents its sum and difference with the tensor trace using the auxiliary coefficients
\begin{equation}
\begin{aligned}
\lambda^{\epsilon}_{1,c}
&=\frac{J_{11,c}+J_{22,c}+\Delta_{i,c}}{2},\\
\lambda^{\epsilon}_{2,c}
&=\frac{J_{11,c}+J_{22,c}-\Delta_{i,c}}{2}.
\end{aligned}
\label{eq:pm_eigenvalues}
\end{equation}
Using these auxiliary coefficients, we compute the
regularized tensor-contrast response as

\begin{equation}
R_{i,c}
=
\frac{
\lambda^{\epsilon}_{1,c}-\lambda^{\epsilon}_{2,c}
}{
\lambda^{\epsilon}_{1,c}+\lambda^{\epsilon}_{2,c}+\epsilon
}
=
\frac{\Delta_{i,c}}
{J_{11,c}+J_{22,c}+\epsilon}.
\label{eq:pm_coherence}
\end{equation}

where $\epsilon=10^{-6}$ is a fixed regularization constant.
The response $R_{i,c}$ characterizes the regularized structural
contrast relative to the local gradient energy. Unlike normalized
coherence measures, it is not strictly bounded within $[0,1]$,
and its magnitude may be influenced by regularization in
low-gradient regions. Therefore, it is employed as an auxiliary
structural descriptor for learned feature modulation.
The channel-wise responses are stacked into
$\mathbf R_i\in\mathbb R^{C_i\times H_i\times W_i}$
and combined with the MSI features to generate the modulation map:
\begin{equation}
\begin{aligned}
\mathbf A_i^S
&=
\sigma
\left[
\operatorname{DWConv}_{3\times3}
\left(
\mathbf S_i
\odot
\mathbf R_i
\right)
\right],\\
\widetilde{\mathbf S}_i
&=
\mathbf S_i
+
\operatorname{Conv}_{1\times1}
\left(
\mathbf S_i
\odot
\mathbf A_i^S
\right).
\end{aligned}
\label{eq:pm_structure_guidance}
\end{equation}
Here, $\sigma$ denotes Sigmoid and $\odot$ denotes element-wise
multiplication. $\operatorname{DWConv}_{3\times3}$ and
$\operatorname{Conv}_{1\times1}$ denote depthwise and pointwise
convolutions, respectively.
$\mathbf A_i^S\in(0,1)^{C_i\times H_i\times W_i}$
is the learned modulation map, and $\widetilde{\mathbf S}_i$
is the resulting MSI guidance representation.

\subsubsection{Low-Frequency-Guided High-Frequency Refinement}

MSI structural information alone cannot completely determine how the HSI feature should be refined because the required update also depends on the spatial details already represented by the current HSI feature. We therefore explicitly separate the HSI representation into local coarse and detail components:
\begin{equation}
\begin{aligned}
\mathbf E_i^{\mathrm{low}}
&=
\mathcal P_5(\mathbf E_i),\\
\mathbf E_i^{\mathrm{high}}
&=
\mathbf E_i
-
\mathbf E_i^{\mathrm{low}},
\end{aligned}
\label{eq:pm_detail_decomposition}
\end{equation}
where $\mathcal P_5$ denotes $5\times5$ average pooling
with unit stride and padding two.
$\mathbf E_i^{\mathrm{low}}$ captures the local average,
while $\mathbf E_i^{\mathrm{high}}$ represents the residual
spatial details.

The low-frequency component captures the surrounding reconstruction context and is used to regulate the high-frequency residual:
\begin{equation}
\begin{aligned}
\mathbf A_i^E
&=
\sigma\left[
\eta_2\left(
\delta\left(
\eta_1(\mathbf E_i^{\mathrm{low}})
\right)
\right)
\right],\\
\widetilde{\mathbf E}_i^{\mathrm{high}}
&=
\mathbf E_i^{\mathrm{high}}
\odot \mathbf A_i^E,
\end{aligned}
\label{eq:pm_detail_modulation}
\end{equation}
where $\eta_1(\cdot)$ and $\eta_2(\cdot)$ are channel-preserving $1\times1$ convolutions and $\delta(\cdot)$ denotes ReLU. The resulting $\widetilde{\mathbf E}_i^{\mathrm{high}}$ represents HSI detail responses modulated according to their surrounding coarse context.

The two conditioning representations provide different information: $\widetilde{\mathbf S}_i$ carries the MSI features and their tensor-based modulation, whereas $\widetilde{\mathbf E}_i^{\mathrm{high}}$ carries HSI detail responses weighted by local coarse context. They are jointly supplied to the kernel predictor.

\subsubsection{Structure-Guided Dynamic Kernel Aggregation}

To adapt local aggregation to spatial structures, we predict
a dynamic neighborhood kernel for each feature channel and
spatial location. The structure-encoded MSI feature
$\widetilde{\mathbf S}_i$ and the context-modulated HSI detail
feature $\widetilde{\mathbf E}_i^{\mathrm{high}}$ jointly
condition the kernel predictor:
\begin{equation}
\mathbf K_i
=
\mathcal K\left(
[\widetilde{\mathbf E}_i^{\mathrm{high}},
\widetilde{\mathbf S}_i]
\right),
\label{eq:pm_spatial_kernel}
\end{equation}
where $\mathcal K$ combines pointwise feature modulation
and kernel prediction. The modulation stage uses two
$1\times1$ convolutions with an intermediate ReLU and
a final Sigmoid. The prediction stage uses two further
$1\times1$ convolutions with an intermediate ReLU and
a linear output. The resulting coefficients are reshaped
into $\mathbf K_i\in
\mathbb R^{C_i\times k^2\times H_i\times W_i}$,
providing a $k\times k$ kernel at each channel and location,
with $k=3$.

The predicted kernels aggregate neighboring HSI features
to produce a residual update:
\begin{equation}
\widetilde E_i(c,p)
=
E_i(c,p)
+
\sum_{r\in\Omega_k}
K_i(c,r,p)\,E_i(c,p+r),
\label{eq:pm_neighborhood_aggregation}
\end{equation}
where $c$ indexes feature channels, $p$ denotes a
two-dimensional spatial location, and $\Omega_k$ contains
the $k^2$ neighborhood offsets. The coefficient
$K_i(c,r,p)$ weights the HSI feature $E_i(c,p+r)$ at
offset $r$ from location $p$. Locations outside the
feature boundary are zero padded.

The MSI structure and HSI detail features determine
the aggregation weights, while $\mathbf E_i$ supplies
the features being aggregated. The refined representation
$\widetilde{\mathbf E}_i$ is then passed to subsequent
fusion stages.

\subsection{Spatial--Spectral Frequency Calibration Attention}
\label{subsec:ssfca}

Following structure-adaptive HSI refinement, Spatial--Spectral Frequency Calibration Attention (SSFCA) performs frequency-aware dependency learning in the progressive fusion branch. As illustrated in Fig.~\ref{fig:overall_architecture}, SSFCA is embedded in a Spatial--Spectral Frequency Calibration Transformer Layer (SSFCTL), which consists of two batch normalization operations, an SSFCA module, a feed-forward network (FFN), and two residual connections. The initial fusion stage operates on shallow MSI and HSI representations, whereas subsequent stages receive the corresponding MSI features, SC-DFF-refined HSI features, and preceding fused representations. Within SSFCA, two parallel branches construct dependencies along spatial-token and learned latent-channel dimensions, respectively. Their outputs are subsequently integrated through a shared branch-calibration mechanism.

For an input feature map $\mathbf U\in\mathbb R^{C\times H_s\times W_s}$, the SSFCTL performs the following residual updates:
\begin{equation}
\begin{aligned}
\mathbf U'
&=\mathbf U+\mathcal W^{-1}\!\left[
\operatorname{SSFCA}\!\left(
\mathcal W\!\left(\operatorname{BN}_1(\mathbf U)\right)
\right)\right],\\
\mathbf U_{\mathrm{out}}
&=\mathbf U'+\operatorname{FFN}\!\left(
\operatorname{BN}_2(\mathbf U')\right),
\end{aligned}
\label{eq:pm_feature_update}
\end{equation}
where $\operatorname{BN}_1$ and $\operatorname{BN}_2$ denote batch normalization, $\mathcal W$ partitions the feature map into nonoverlapping local windows, and $\mathcal W^{-1}$ restores their spatial arrangement. The FFN employs pointwise channel expansion, $3\times3$ depthwise convolution, and GELU-gated feature interaction. Both residual updates preserve the input feature dimensions. For each window, let $\mathbf T\in\mathbb R^{N\times C}$ denote its flattened representation, where $N=M^2$ for a window of size $M\times M$. The $C$ channels are divided into $n_h$ attention heads, each with $d=C/n_h$ latent channels. Batch and window indices are omitted below for clarity. The following sections introduce the latent spectral-frequency and spatial-token frequency attention branches before describing their complementary fusion.

\subsubsection{Latent Spectral-Frequency Attention}
\begin{figure}[t]
    \centering   
    \includegraphics[width=0.98\columnwidth]{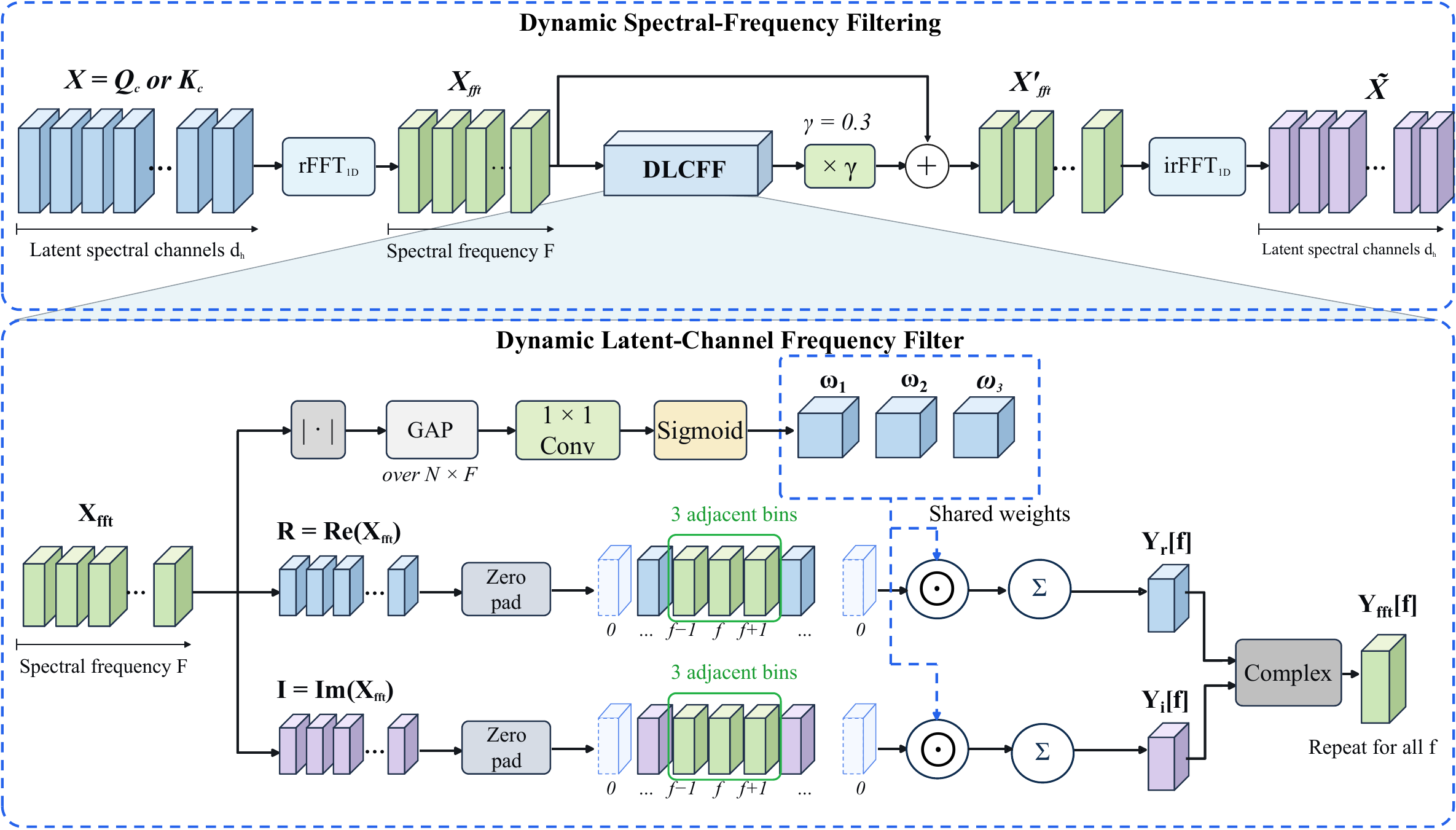}
        \caption{Dynamic Spectral--Frequency Filtering (DSFF) within SSFCA. fourier-transformed latent-channel queries and keys undergo residual frequency calibration. Its core Dynamic Latent-Channel Frequency Filter module uses pooled frequency magnitudes to generate three-tap kernels that aggregate neighboring frequency bins with shared weights for real and imaginary components, followed by inverse transformation.}
    \label{fig:dynamic_spectral_filter}
    \vspace{-6pt}
\end{figure}

The latent spectral-frequency branch incorporates Dynamic Spectral--Frequency Filtering (DSFF) to calibrate projected query and key representations before constructing channel affinities. As shown in Fig.~\ref{fig:dynamic_spectral_filter}, DSFF transforms these representations along the latent-channel dimension and applies input-conditioned three-tap filtering to neighboring frequency bins. Here, the latent channels are learned feature coordinates rather than physical HSI wavelengths; thus, the filtering captures dependencies in the projected feature space without assuming physical wavelength continuity.
A $1\times1$ pointwise convolution followed by a $3\times3$ depthwise convolution first generates the query, key, and value features $\mathbf Q^c_a,\mathbf K^c_a,\mathbf V^c_a\in\mathbb R^{N\times d}$ for the $a$-th attention head. For either the query or key features, let
$\mathbf X^c_a\in\mathbb R^{N\times d}$ denote the
representation of the $a$-th head. Stacking all heads gives
$\mathbf X^c=\operatorname{Stack}_{a=0}^{n_h-1}
(\mathbf X^c_a)\in\mathbb R^{n_h\times N\times d}$.
A real-input Fourier transform along the last,
latent-channel dimension then yields
\begin{equation}
\boldsymbol\Xi
=\mathcal F_d(\mathbf X^c)
\in\mathbb C^{n_h\times N\times L_d},
\qquad
L_d=\left\lfloor\frac d2\right\rfloor+1,
\label{eq:pm_latent_fft}
\end{equation}
where $\mathcal F_d$ denotes the one-dimensional real-input fourier transform along the last dimension, and $L_d$ is the number of retained one-sided frequency bins.

The core Dynamic Latent-Channel Frequency Filter module uses the fourier magnitudes to generate a content-conditioned three-tap kernel. Specifically, the magnitudes are pooled over spatial tokens and retained frequency bins within each window:
\begin{equation}
\begin{aligned}
z_a
&=\frac{1}{NL_d}
\sum_{p=1}^{N}\sum_{f=0}^{L_d-1}
\left|\Xi_a(p,f)\right|,\\
\mathbf w
&=\operatorname{Reshape}\!\left[
\sigma\!\left(\mathbf W_g\mathbf z+\mathbf b_g\right)
\right],
\end{aligned}
\label{eq:pm_frequency_kernel}
\end{equation}
where $p$ and $f$ index the spatial tokens and
retained latent-frequency bins, respectively,
$\mathbf z\in\mathbb R^{n_h}$ contains the
head-wise magnitude descriptors,
$\mathbf W_g\in\mathbb R^{3n_h\times n_h}$ and
$\mathbf b_g$ denote the parameters of a pointwise kernel
generator, and $\sigma(\cdot)$ denotes Sigmoid.
The generated weights
$\mathbf w\in(0,1)^{n_h\times3}$ vary with the input window
and attention head but are shared across token and frequency
positions within each head. The pointwise kernel generator can
mix information across different heads.

For each head, the three-tap kernel aggregates adjacent frequency bins as
\begin{equation}
\left[\mathcal D(\boldsymbol\Xi)\right]_a(p,f)
=\sum_{r=-1}^{1}w_{a,r}\,\Xi_a(p,f+r),
\label{eq:pm_frequency_filtering}
\end{equation}
where out-of-range frequency positions are zero padded.
Since $\boldsymbol\Xi$ is complex-valued and
$w_{a,r}$ is real-valued, the same three-tap kernel
is applied to the real and imaginary components
independently, as illustrated in
Fig.~\ref{fig:dynamic_spectral_filter}.
Residual frequency calibration and inverse
transformation then produce
\begin{equation}
\widetilde{\mathbf X}^{c}
=\mathcal F_d^{-1}\!\left[
\boldsymbol\Xi+\gamma\mathcal D(\boldsymbol\Xi)
\right],
\label{eq:pm_frequency_residual}
\end{equation}
where $\gamma=0.3$ and the inverse real fourier transform reconstructs $d$ latent channels. This operation changes the frequency representation before attention estimation; it does not impose explicit phase preservation. Although queries and keys share the Dynamic Latent-Channel Frequency Filter parameters, their filtering weights are independently conditioned on their respective fourier magnitudes.

Following inverse transformation, the calibrated queries and keys are $\ell_2$-normalized along the latent-channel dimension, yielding $\overline{\mathbf Q}^c_a$ and $\overline{\mathbf K}^c_a$. Latent-channel affinities and their aggregated responses are then computed as
\begin{equation}
\begin{aligned}
\mathbf A^c_a
&=\operatorname{Softmax}\!\left[
\tau_a(\overline{\mathbf Q}^c_a)^\top
\overline{\mathbf K}^c_a+\mathbf B^c_a
\right],\\
\mathbf O^c_a
&=\left[\mathbf A^c_a(\mathbf V^c_a)^\top\right]^\top,\\
\mathbf O_c
&=\operatorname{Proj}_c\!\left[
\operatorname{Concat}_a(\mathbf O^c_a)
\right],
\end{aligned}
\label{eq:pm_channel_output}
\end{equation}
where $\mathbf A^c_a,\mathbf B^c_a\in\mathbb R^{d\times d}$, $\tau_a$ is a learnable head-wise temperature shared with the spatial-token branch, $\operatorname{Concat}_a$ concatenates attention-head outputs, and $\operatorname{Proj}_c$ is a $1\times1$ output projection. Softmax acts along the final affinity dimension.

A fixed log-decay positional bias is included to encode relative distances along the learned latent-channel coordinates:
\begin{equation}
B^c_a(u,v)
=|u-v|\log\!\left(
1-2^{-(2+4a/n_h)}
\right),
\label{eq:pm_channel_bias}
\end{equation}
where $a=0,\ldots,n_h-1$ and $u,v\in\{1,\ldots,d\}$. This bias favors neighboring indices in the learned feature space without implying that adjacent latent channels correspond to adjacent physical wavelengths.

\subsubsection{Spatial-Token Frequency Attention}

The spatial-token frequency branch introduces fourier-domain interaction into the query--key representations before spatial affinity estimation. For the $a$-th attention head, linear projections produce $\mathbf Q^s_a,\mathbf K^s_a,\mathbf V^s_a\in\mathbb R^{N\times d}$. The query and key features undergo a one-dimensional real-input fourier transform along the flattened spatial-token dimension, followed by head-wise scaling and element-wise frequency multiplication:
\begin{equation}
\begin{aligned}
\widehat{\mathbf Q}^s_a
&=\omega_a\mathcal F_N(\mathbf Q^s_a),\\
\widehat{\mathbf K}^s_a
&=\omega_a\mathcal F_N(\mathbf K^s_a),\\
\mathbf P_a
&=\mathcal F_N^{-1}\!\left(
\widehat{\mathbf Q}^s_a\odot
\widehat{\mathbf K}^s_a
\right),
\end{aligned}
\label{eq:pm_spatial_interaction}
\end{equation}
where $\mathcal F_N$ and $\mathcal F_N^{-1}$ are the one-dimensional real-input fourier transform and its inverse along the $N$ tokens, respectively, $\odot$ denotes element-wise multiplication, and $\omega_a$ is a learnable real-valued scale initialized to one. The same $\omega_a$ is used for all token-frequency bins and projected channels within a head, so it acts as a head-wise gain rather than a frequency-selective filter. The resulting $\mathbf P_a\in\mathbb R^{N\times d}$ preserves the original token and channel dimensions.

The interaction is performed along the serialized token dimension within each local window, providing frequency-domain mixing of the projected query and key representations while preserving their original token and channel dimensions.

Spatial-token affinities are subsequently obtained from the fourier-interacted representation:
\begin{equation}
\begin{aligned}
\mathbf A^s_a
&=\operatorname{Softmax}\!\left(
\frac{\tau_a}{\sqrt d}\,
\mathbf P_a\mathbf P_a^\top+\mathbf B^s_a
\right),\\
\mathbf O_s
&=\operatorname{Proj}_s\!\left[
\operatorname{Concat}_a\!\left(
\mathbf A^s_a\mathbf V^s_a
\right)\right],
\end{aligned}
\label{eq:pm_spatial_output}
\end{equation}
where $\mathbf A^s_a,\mathbf B^s_a\in\mathbb R^{N\times N}$, $\mathbf B^s_a$ denotes a learned two-dimensional relative positional bias, and $\operatorname{Proj}_s$ denotes the output projection. Here, $\mathbf{P}_a \mathbf{P}_a^\top$ constructs pairwise affinities between the fourier-interacted token representations rather than a direct directional query-key matching matrix.
The positional bias supplies two-dimensional relative-position information after the one-dimensional token-frequency interaction. Because the $N\times N$ affinity matrix is explicitly computed, the operation retains quadratic token-affinity complexity within each window.

\subsubsection{Complementary Feature Fusion}

The two attention branches operate on the same window features but aggregate responses along complementary dimensions: the spectral-frequency branch models dependencies among projected latent channels, whereas the spatial-frequency branch models relationships among spatial tokens. Their outputs $\mathbf O_c,\mathbf O_s\in\mathbb R^{N\times C}$ are combined as
\begin{equation}
\begin{aligned}
\mathbf T_{\mathrm{out}}
&=\operatorname{SSFCA}(\mathbf T)\\
&=\mathbf O_s+(1+\mathbf g)\odot\mathbf O_c,
\end{aligned}
\label{eq:pm_ssfca_output}
\end{equation}
where $\boldsymbol{g} \in (0,1)^{N\times 1}$ denotes a broadcast shared calibration factor applied to the latent-channel response.
The spatial response is added directly, while the latent-channel response is scaled by a residual calibration factor.

Specifically, the normalized spatial-attention responses are pooled along the attention-token dimension and passed through two pointwise projections, batch normalization, and Sigmoid to obtain the shared calibration factor:
\begin{equation}
\begin{aligned}
\mu_a(p)
&=\frac{1}{N}\sum_{q=1}^{N}A^s_a(p,q),\\
\mathbf g
&=\sigma\!\left[
\psi_2\!\left(
\operatorname{BN}\!\left(
\psi_1(\boldsymbol\mu)
\right)\right)\right],
\end{aligned}
\label{eq:pm_branch_scale}
\end{equation}
where $\mu$ denotes the row-wise averaged attention responses across all heads, and $\psi_1(\cdot)$ and $\psi_2(\cdot)$ are pointwise projections. Since each row of the spatial attention matrix is normalized by Softmax, $\mu_a(p)=1/N$. Therefore, the calibration branch produces a spatially shared, input-independent gain that balances the contributions of the spatial-token and latent-channel branches.

Since the spatial affinity has already been normalized by Softmax, the pooled response serves as a normalized reference for the subsequent learnable branch calibration. Accordingly, $\boldsymbol{g}$ is interpreted as a learned shared branch gain rather than a content-dependent spatial gate. The input-dependent modeling of SSFCA is provided by the spatial-token and latent-channel affinities together with the dynamic spectral-frequency filtering. Through the residual updates in Eq.~\eqref{eq:pm_feature_update}, the resulting frequency-aware representation is propagated to subsequent fusion stages for multiscale HR-HSI reconstruction.

\section{Experiments}

\subsection{Experimental Settings}

\subsubsection{Datasets}
Three hyperspectral datasets, including CAVE \cite{yasuma2010generalized}, NTIRE2022 \cite{arad2022ntire}, and Gaofen-5 (GF5) \cite{liu2025low},
are used to evaluate the proposed FSCM-Net. The CAVE dataset contains
32 HSIs with a spatial size of $512\times512$ and 31 spectral bands
ranging from 400 to 700 nm, where the first 20 HSIs are used for
training and the remaining 12 for testing. 
The NTIRE2022 dataset  contains HSIs of size
$512\times482\times31$, covering the spectral range of 400--700 nm.
The 900 training images are used for training, and the 50 validation
images are used for evaluation. For real-world evaluation, we employ the Gaofen5 (GF5) dataset , comprising a 150-band HSI acquired by the Chinese Gaofen-5 satellite and a four-band MSI captured by the Gaofen-1 satellite. The HSI and MSI have spatial dimensions of $1100 \times 1144$ and $2200 \times 2288$, respectively, corresponding to a spatial resolution ratio of 2. Since no ground-truth HR-HSI is available, the dataset is used to evaluate fusion performance under real-world imaging conditions. The statistics of the three datasets are summarized in
Table~\ref{tab:dataset_statistics}.

\suppressfloats[t]

\begin{table}[!t]
\caption{Statistics of the Datasets Used in the Experiments}
\label{tab:dataset_statistics}
\centering
\footnotesize
\setlength{\tabcolsep}{2.5pt}
\renewcommand{\arraystretch}{1.12}

\begin{tabular*}{\columnwidth}
{@{\extracolsep{\fill}}lccc@{}}
\toprule
Dataset & Dimensions & Samples & Range (nm) \\
\midrule

CAVE~\cite{yasuma2010generalized}
& $512\times512\times31$
& 32
& 400--700 \\

NTIRE2022~\cite{arad2022ntire}
& $512\times482\times31$
& 950
& 400--700 \\

GF5 (HSI)
& $1100\times1144\times150$
& 1 pair
& -- \\

GF1 (MSI)
& $2200\times2288\times4$
&
& -- \\

\bottomrule
\end{tabular*}
\end{table}

\subsubsection{Data Simulation}
For CAVE and NTIRE2022, the original HSI is regarded as the reference HR-HSI, while the corresponding LR-HSI and HR-MSI are generated through spatial and spectral degradation, respectively. For both datasets, an $8\times8$ Gaussian filter with a standard deviation of 3 is applied to the HR-HSI, followed by an $8\times$ spatial downsampling to generate the LR-HSI. The HR-MSI is generated using the spectral response function of the Nikon D700 camera. During training, the HR-HSIs are cropped into $64\times64$ patches, corresponding to LR-HSI patches of $8\times8\times31$ and HR-MSI patches of $64\times64\times3$. For NTIRE2022, reflection padding is additionally applied before generating the training samples.

For the real-world GF5 dataset, Gaussian blurring and spatial downsampling with a scale factor of 2 are applied separately to the observed HSI and MSI to construct reduced-resolution training pairs. Specifically, the downsampled MSI and original HSI are cropped into $64 \times 64$ patches with a stride of 32, while the downsampled HSI is cropped into $32 \times 32$ patches with a stride of 16. The reduced-resolution HSI and MSI patches serve as network inputs, with the corresponding original HSI patches used as supervision targets. During testing, the original HSI and MSI observations are directly fed into the trained network to reconstruct the full-resolution HSI.

\subsubsection{Implementation Details}
All experiments were conducted using NVIDIA GeForce RTX 4090 GPUs.
The network was implemented in PyTorch and optimized using Adam with
$\beta_1=0.9$, $\beta_2=0.999$, and a weight decay of $1\times10^{-8}$.
The mean $\ell_1$ reconstruction loss was adopted as the training objective.
For CAVE, NTIRE2022, and GF5, the initial learning rate was set
to $4\times10^{-4}$. A cosine annealing schedule was applied after
each optimization iteration, gradually reducing the learning rate
to $1\times10^{-6}$. For CAVE and NTIRE2022, the batch size was
set to 16, and training patches of $64\times64$ were extracted
with a stride of 32. Full-image reconstruction on these two datasets
was performed using a $64\times64$ sliding window with the same stride.

All convolutional and linear layers are initialized using Xavier uniform
initialization. The base feature width of FSCM-Net is 32. The three
SSFCA-based transformer stages use feature dimensions of 64, 64, and 128,
with 4, 8, and 16 attention heads, respectively. Their corresponding
window sizes are 8, 4, and 4, and the FFN expansion ratio is set to 2.
Batch normalization is employed before both SSFCA and the FFN. In SSFCA,
the dynamic latent spectral-frequency filter adopts a three-tap kernel
and the residual modulation coefficient is set to $\gamma=0.3$. In
SC-DFF, a $5\times5$ average-pooling operator is used to extract the
local low-frequency context, and a $3\times3$ dynamic aggregation kernel
is adopted by default.
\subsection{Compared Methods}

To comprehensively evaluate the effectiveness of the proposed
FSCM-Net, seven representative state-of-the-art hyperspectral image
fusion methods are selected for comparison, including LGCT \cite{he2024lgct}, AELF\cite{he2025adaptive},
OTIAS\cite{deng2025otias}, HSC\cite{liu2025heterospectral}, LRTN\cite{liu2025low}, SSDAN\cite{song2026s}, and CLSNet\cite{yang2026correlation}. These methods cover recent
representative fusion architectures with different spatial--spectral
feature modeling strategies. For a fair comparison, all methods are
evaluated using the same training and testing data, and the
corresponding experimental settings follow their original papers or
publicly available implementations.

\subsection{Quality Metrics}

For CAVE and NTIRE2022, five full-reference metrics are
adopted to evaluate reconstruction quality: peak signal-to-noise
ratio (PSNR), spectral angle mapper (SAM), erreur relative
globale adimensionnelle de synth\`ese (ERGAS), structural
similarity index measure (SSIM), and root mean square error
(RMSE). PSNR and RMSE quantify reconstruction errors,
SAM measures spectral angular discrepancies, ERGAS assesses
relative reconstruction errors, and SSIM evaluates structural
similarity. Higher PSNR and SSIM and lower SAM, ERGAS,
and RMSE indicate better performance.

For the real-world GF5 dataset, ground-truth HR-HSI is
unavailable. Therefore, the quality with no reference (QNR)
metric \cite{alparone2008multispectral} is employed to assess
full-resolution fusion quality through spectral and spatial
distortion measures. A higher QNR value indicates better
spatial--spectral consistency under the adopted criterion.

\subsection{Experimental Results on Simulated Datasets}
\begin{table}[!t]
\caption{Quantitative Comparison on the CAVE Dataset}
\label{tab:cave_results}
\centering
\footnotesize
\setlength{\tabcolsep}{1.5pt}
\renewcommand{\arraystretch}{1.12}

\begin{tabular*}{\columnwidth}{@{\extracolsep{\fill}}lccccc}
\toprule
Method &
PSNR$\uparrow$ &
SAM$\downarrow$ &
ERGAS$\downarrow$ &
SSIM$\uparrow$ &
RMSE$\downarrow$ \\
\midrule

LGCT \cite{he2024lgct}
& 47.6230 & 2.5857 & 0.5928 & 0.9936 & 1.3436 \\

AELF \cite{he2025adaptive}
& 47.9780 & 2.4287 & 0.5629 & 0.9943 & 1.2856 \\

OTIAS \cite{deng2025otias}
& 47.8690 & 2.3799 & 0.5681 & 0.9944 & 1.3192 \\

HSC \cite{liu2025heterospectral}
& 48.7350 & 2.2650 & 0.5221 & 0.9948 & 1.1796 \\

LRTN \cite{liu2025low}
& 47.9370 & 2.3322 & 0.5661 & 0.9943 & 1.3109 \\

SSDAN \cite{song2026s}
& \underline{48.8550}
& \underline{2.2497}
& \underline{0.5112}
& \underline{0.9949}
& \underline{1.1664} \\

CLSNet \cite{yang2026correlation}
& 47.7466 & 2.7329 & 0.9558 & 0.9856 & 1.6115 \\

FSCM-Net (Ours)
& \textbf{49.0110}
& \textbf{2.2287}
& \textbf{0.5037}
& \textbf{0.9952}
& \textbf{1.1341} \\

\bottomrule
\end{tabular*}
\end{table}
\begin{table}[!t]
\caption{Quantitative Comparison on the NTIRE2022 Dataset}
\label{tab:ntire_results}
\centering
\footnotesize
\setlength{\tabcolsep}{1.5pt}
\renewcommand{\arraystretch}{1.12}
\begin{tabular*}{\columnwidth}{@{\extracolsep{\fill}}lccccc}
\toprule
Method &
PSNR$\uparrow$ &
SAM$\downarrow$ &
ERGAS$\downarrow$ &
SSIM$\uparrow$ &
RMSE$\downarrow$ \\
\midrule
LGCT \cite{he2024lgct}
& 52.7499 & 1.3611 & 0.7008 & 0.9969 & 0.8566 \\
AELF \cite{he2025adaptive}
& 53.2158 & 1.1888 & 0.5829 & 0.9978 & 0.7747 \\
OTIAS \cite{deng2025otias}
& 54.4937 & 1.0456 & \textbf{0.5317} & \underline{0.9979} & 0.7376 \\
HSC \cite{liu2025heterospectral}
& 55.0173 & 1.0477 & 0.6500 & 0.9975 & 0.7203 \\
LRTN \cite{liu2025low}
& 54.1834 & 1.0701 & 0.6695 & 0.9970 & 0.7959 \\
SSDAN \cite{song2026s}
& \underline{55.1088}
& \underline{0.9847}
& \underline{0.5425}
& 0.9977
& \underline{0.6839} \\
CLSNet \cite{yang2026correlation}
& 54.8098 & 1.0490 & 0.6430 & 0.9976 & 0.7743 \\
FSCM-Net (Ours)
& \textbf{55.4190}
& \textbf{0.9634}
& 0.5814
& \textbf{0.9980}
& \textbf{0.6737} \\
\bottomrule
\end{tabular*}
\end{table}

\begin{figure*}[!t]
    \centering
    \includegraphics[width=0.99\textwidth]{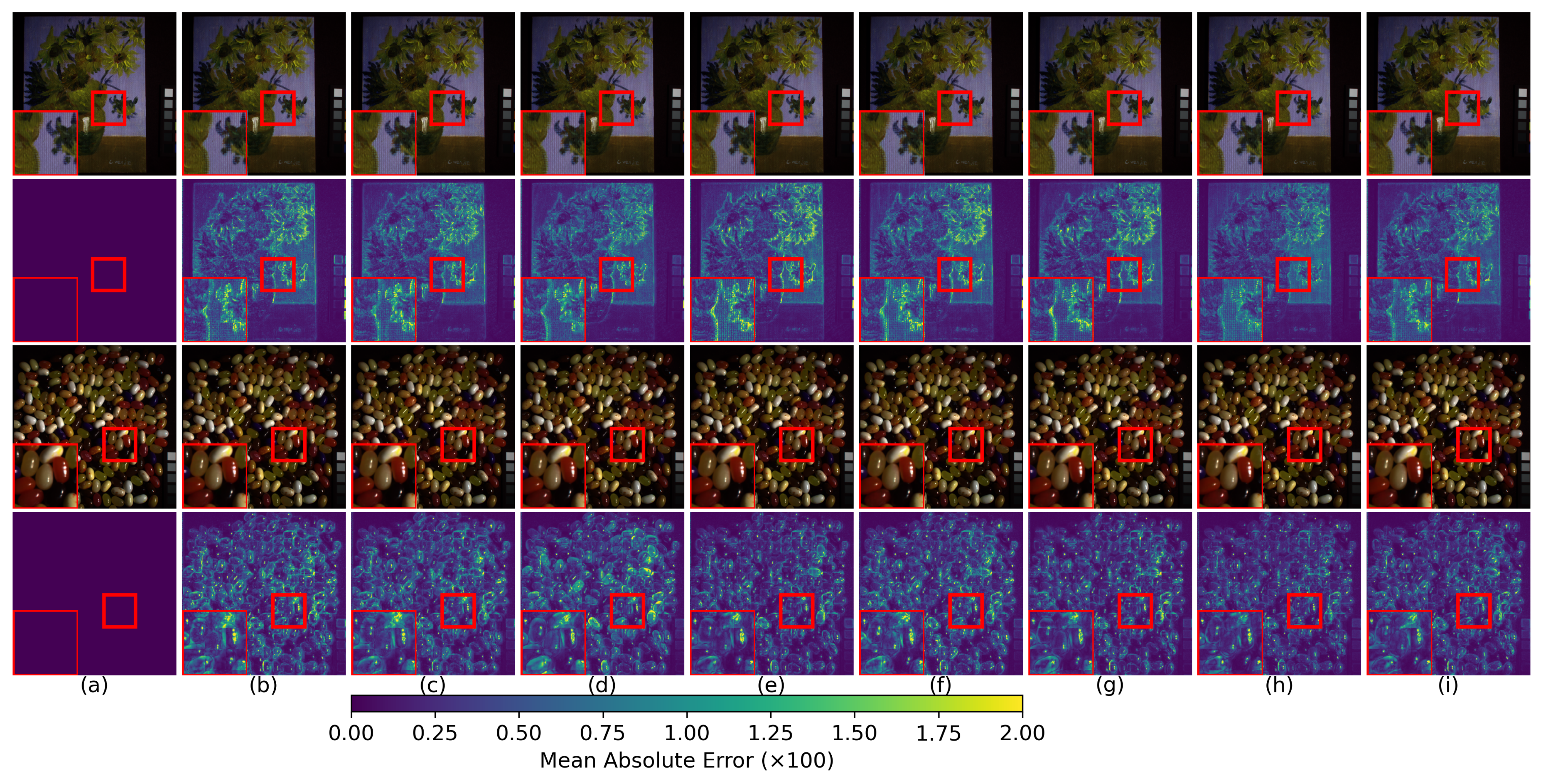}
    \caption{Qualitative comparison of different methods on the CAVE
    dataset. The first and third rows show the reconstructed images,
    while the second and fourth rows show the corresponding error maps.
    The red boxes highlight enlarged local regions. From left to right:
    (a) GT, (b) LGCT, (c) AELF, (d) OTIAS, (e) HSC, (f) LRTN,
    (g) SSDAN, (h) CLSNet, and (i) FSCM-Net (Ours).}
    \label{fig:cave_visual}
\end{figure*}
\begin{figure*}[!t]
\centering
\includegraphics[width=\linewidth]{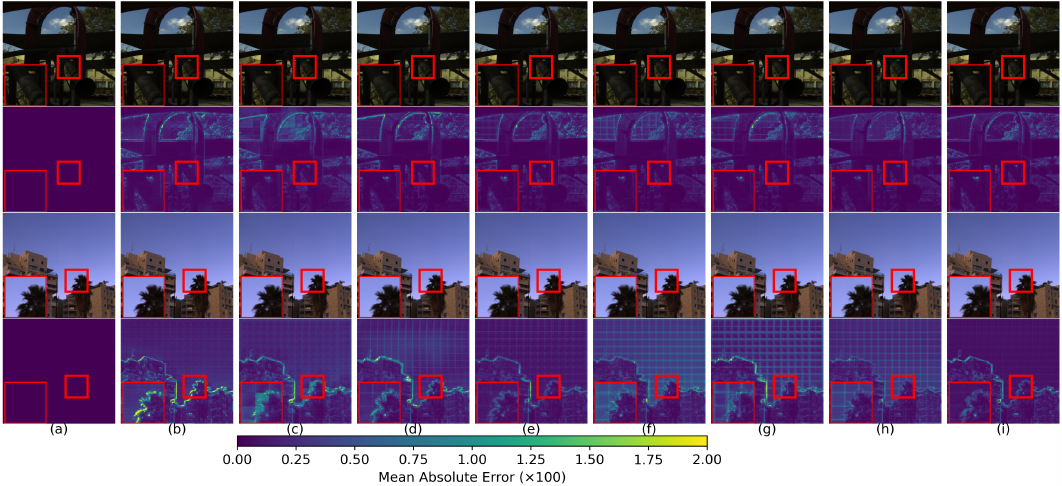}
\caption{Visual comparison of the reconstructed HR-HSIs and the
corresponding error maps on two representative scenes from the
NTIRE2022 dataset. (a) Reference. (b)--(i) Results of LGCT, AELF,
OTIAS, HSC, LRTN, SSDAN, CLSNet, and FSCM-Net (Ours), respectively.
Red rectangles highlight the regions of interest for close-up
inspection, and the color bar indicates the magnitude of the
reconstruction error.}
\label{fig:ntire2022_visual}
\end{figure*}
\subsubsection{Quantitative Comparison}
Table~\ref{tab:cave_results} presents the quantitative comparison
on the CAVE test set. FSCM-Net achieves the best performance across
all five metrics, indicating favorable reconstruction accuracy,
spectral fidelity, and structural preservation. Specifically, it
obtains a PSNR of 49.0110~dB, exceeding the second-best SSDAN by
0.1560~dB, while reducing RMSE from 1.1664 to 1.1341. These results
indicate lower pixel-wise deviations from the reference HR-HSIs.
Moreover, the lowest SAM of 2.2287 demonstrates better preservation
of spectral shapes, while the highest SSIM of 0.9952 indicates
closer structural agreement with the reference images. The lowest
ERGAS of 0.5037 further supports its reconstruction quality in
terms of relative errors.

Table~\ref{tab:ntire_results} reports the results on NTIRE2022.
FSCM-Net achieves the best PSNR, SAM, SSIM, and RMSE among the
compared methods. Its PSNR reaches 55.4190~dB, improving upon
SSDAN by 0.3102~dB, with RMSE decreasing from 0.6839 to 0.6737.
The improved reconstruction accuracy is accompanied by a lower
SAM of 0.9634 and a higher SSIM of 0.9980, showing that the
reduction in reconstruction errors is achieved while preserving
spectral shapes and spatial structures. Although OTIAS obtains
a lower ERGAS, FSCM-Net provides favorable performance across
the other four metrics.

\subsubsection{Qualitative Comparison}
Fig.~\ref{fig:cave_visual} presents the reconstructed images and
corresponding error maps for two CAVE scenes. The enlarged regions
show that FSCM-Net preserves fine plant structures and boundaries
between adjacent beads, producing results visually close to the
reference images. Its error maps exhibit generally weaker responses
in the selected regions, indicating reduced local reconstruction
errors. These observations complement the quantitative improvements
in reconstruction accuracy and structural consistency.

Fig.~\ref{fig:ntire2022_visual} shows the visual comparison on
NTIRE2022. FSCM-Net retains clear vehicle contours and building
boundaries in the highlighted regions, with limited visible
blurring. The corresponding error maps also show relatively low
responses around these structures. Together with the PSNR and
SSIM results, these observations support its ability to reconstruct
spatial details while maintaining overall image consistency.

\subsection{Experimental Results on the Real Dataset}
\begin{table*}[!t]
\caption{No-reference quality assessment (QNR) on the GF5 dataset. The best result is in bold and the second best is underlined.}
\label{tab:gf5_results}
\centering
\small
\setlength{\tabcolsep}{5pt}
\renewcommand{\arraystretch}{1.2}
\begin{tabular}{lcccccccc}
\toprule
Method &
LGCT \cite{he2024lgct} &
AELF \cite{he2025adaptive} &
OTIAS \cite{deng2025otias} &
HSC \cite{liu2025heterospectral} &
LRTN \cite{liu2025low} &
SSDAN \cite{song2026s} &
CLSNet \cite{yang2026correlation} &
FSCM-Net (Ours) \\
\midrule
\addlinespace
QNR$\uparrow$ &
0.9798 &
0.9803 &
\underline{0.9804} &
0.9761 &
0.9781 &
\underline{0.9804} &
0.9789 &
\textbf{0.9856} \\
\bottomrule
\end{tabular}
\end{table*}
\begin{figure*}[!t]
\centering
\includegraphics[width=\linewidth]{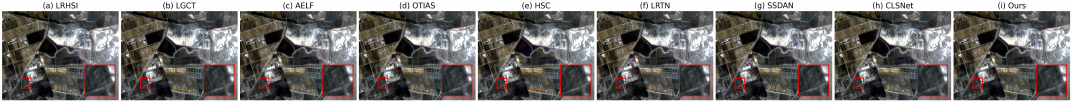}
\caption{Visual comparison of the full-resolution fusion results on
the GF5 dataset. (a) LRHSI. (b)--(i) Results of LGCT, AELF,
OTIAS, HSC, LRTN, SSDAN, CLSNet, and FSCM-Net (Ours), respectively.
Red rectangles highlight the regions of interest for close-up
inspection.}
\label{fig:gf5_visual}
\end{figure*}
\subsubsection{Quantitative Comparison}
Since GF5 does not provide ground-truth HR-HSIs, QNR is used
to assess full-resolution fusion quality. As reported in
Table~\ref{tab:gf5_results}, FSCM-Net achieves the highest QNR
of 0.9856, exceeding the joint second-best value of 0.9804
obtained by OTIAS and SSDAN. This result indicates better no-reference spatial--spectral consistency under the adopted QNR criterion, supporting the practical applicability of FSCM-Net to the evaluated real remote-sensing scene.

\subsubsection{Qualitative Comparison}
Fig.~\ref{fig:gf5_visual} presents the full-resolution fusion
results on GF5. In the highlighted region, FSCM-Net preserves
building boundaries, road structures, and local textures.
Compared with several competing methods, its reconstruction
exhibits less visible smoothing and more distinct spatial details.
These visual observations complement the QNR results and support
the applicability of FSCM-Net under practical imaging conditions.
\begin{table}[!t]
\caption{Ablation Study of SC-DFF and SSFCA on the CAVE Dataset}
\label{tab:ablation}
\centering
\footnotesize
\setlength{\tabcolsep}{1.5pt}
\renewcommand{\arraystretch}{1.12}

\begin{tabular*}{\columnwidth}{@{\extracolsep{\fill}}lccccc@{}}
\toprule
Method &
PSNR$\uparrow$ &
RMSE$\downarrow$ &
SAM$\downarrow$ &
SSIM$\uparrow$ &
ERGAS$\downarrow$ \\
\midrule

w/o All
& 48.0572
& 1.2718
& 2.4292
& 0.9943
& 0.5622 \\

w/o SC-DFF
& \underline{48.7210}
& \underline{1.1804}
& \underline{2.2899}
& \underline{0.9949}
& \underline{0.5206} \\

w/o SSFCA
& 48.2538
& 1.2556
& 2.3892
& 0.9943
& 0.5537 \\

FSCM-Net (Ours)
& \textbf{49.0110}
& \textbf{1.1341}
& \textbf{2.2287}
& \textbf{0.9952}
& \textbf{0.5037} \\

\bottomrule
\end{tabular*}
\end{table}
\subsection{Ablation Study}
\subsubsection{Contributions of SC-DFF and SSFCA}
To evaluate the contributions of the two main components in FSCM-Net,
ablation experiments are conducted on the CAVE dataset. Specifically,
``w/o SC-DFF'' denotes the variant in which the Structure-Coherent
Dynamic Feature Fusion module is removed, ``w/o SSFCA'' denotes the
variant without Spatial--Spectral Frequency Calibration Attention, and
``w/o All'' removes both components.

As reported in Table~\ref{tab:ablation}, the complete FSCM-Net
achieves the best performance across all five metrics, while
removing either component reduces reconstruction quality. Without SC-DFF, PSNR decreases by 0.2900~dB, accompanied by
increases in RMSE and SAM. The lower reconstruction accuracy
and larger spectral angular errors support the contribution of
structure-adaptive local aggregation to accurate HSI reconstruction. The decrease in SSIM from 0.9952 to 0.9949 further indicates
that SC-DFF helps preserve spatial structures.

Removing SSFCA causes a larger PSNR decrease of 0.7572~dB,
while SAM increases from 2.2287 to 2.3892 and SSIM decreases
to 0.9943. These changes indicate that SSFCA contributes to
both spectral fidelity and structural consistency, supporting
the effectiveness of frequency-calibrated spatial--spectral
interaction. Its removal produces a larger performance drop
than removing SC-DFF under the evaluated configuration. When both components are removed, PSNR decreases to 48.0572~dB,
with the highest RMSE, SAM, and ERGAS among the variants.
The complete model improves PSNR by 0.9538~dB over this baseline. Moreover, it consistently outperforms both variants retaining
only one component, supporting the complementary contributions
of frequency-aware interaction and structure-adaptive local
refinement to reconstruction quality.

\section{Conclusion}
In this paper, we proposed a Frequency--Structure Complementary Modeling Network (FSCM-Net) for HSI--MSI fusion to address the limitations of conventional feature-domain attention and spatially shared local aggregation. The proposed framework couples frequency-aware dependency learning with structure-adaptive local refinement within a multiscale encoder--decoder. Specifically, SSFCA constructs spatial-token affinities from fourier-interacted query--key representations and latent-channel affinities after input-conditioned frequency filtering along learned channel coordinates. The resulting spatial-token and latent-channel responses are combined through shared branch calibration to enrich spatial--spectral representation learning. Complementarily, SC-DFF combines regularized structure-tensor cues from MSI features with context-modulated HSI detail representations to predict channel- and position-dependent aggregation kernels. These kernels adaptively aggregate local HSI features according to structural variations without directly replacing HSI feature values with MSI information. Through progressive integration across multiple scales, the two modules complement each other in spatial detail reconstruction and spectral preservation. Experiments on CAVE, NTIRE2022, and GF5 demonstrate favorable fusion performance compared with representative methods, while ablation studies validate the individual contributions and complementary roles of SSFCA and SC-DFF.
Future work will investigate the robustness of FSCM-Net under more diverse sensor degradations and cross-modal misregistration, as well as more efficient implementations for large-scale hyperspectral image fusion.

\bibliographystyle{IEEEtran}
\bibliography{reference} 
\end{document}